# Classification performance and reproducibility of GPT-4 omni for information extraction from veterinary electronic health records

**Judit M Wulcan[1], Kevin L Jacques[1], Mary Ann Lee[2], Samantha L Kovacs[1], Nicole Dausend[3], Lauren E Prince[1], Jonatan Wulcan[4], Sina Marsilio[3], Stefan M Keller[1]**

[1]University of California Davis, School of Veterinary Medicine, Department of Pathology, Microbiology and Immunology, Davis, CA, USA

[2]Colorado State University, College of Veterinary Medicine and Biomedical Sciences, James L. Voss Veterinary Teaching Hospital, Fort Collins, CO, USA

[3]University of California Davis, School of Veterinary Medicine, Department of Medicine and Epidemiology, Davis, CA, USA

[4]Independent researcher, Malmoe, Sweden

**\*Correspondence**:
Stefan M Keller
smkeller@ucdavis.edu

**Abstract**

Large language models (LLMs) can extract information from veterinary electronic health records (EHRs), but performance difference between models, the effect of temperature settings, and the influence of text ambiguity have not been previously evaluated. This study addresses these gaps by comparing the performance of GPT-4 omni (GPT-4o) and GPT-3.5 Turbo under different conditions and investigating the relationship between human interobserver agreement and LLM errors. The LLMs and five humans were tasked with identifying six clinical signs associated with Feline chronic enteropathy in 250 EHRs from a veterinary referral hospital. At temperature 0, the performance of GPT-4o compared to the majority opinion of human respondents, achieved 96.9% sensitivity, (interquartile range [IQR] 92.9-99.3%), 97.6% specificity (IQR 96.5-98.5%), 80.7% positive predictive value (IQR 70.8-84.6%), 99.5% negative predictive value (IQR 99.0-99.9%), 84.4% F1 score (IQR 77.3-90.4%), and 96.3% balanced accuracy (IQR 95.0-97.9%). The performance of GPT-4o was significantly better than that of its predecessor, GPT-3.5 Turbo, particularly with respect to sensitivity where GPT-3.5 Turbo only achieved 81.7% (IQR 78.9-84.8%). Adjusting the temperature for GPT-4o did not significantly impact classification performance. GPT-4o demonstrated greater reproducibility than human pairs regardless of temperature, with an average Cohen's kappa of 0.98 (IQR 0.98-0.99) at temperature 0 compared to 0.8 (IQR 0.78-0.81) of humans. Most GPT-4o errors occurred in instances where humans disagreed (35/43 errors [81.4%]), suggesting that these errors were more likely caused by ambiguity of the EHR than explicit model faults. Using GPT-4o to automate information extraction from veterinary EHRs is a viable alternative to manual extraction.

# 1 Introduction

Efficient, accurate, and scalable methods for extracting information from electronic health records (EHRs) are essential for conducting retrospective studies in veterinary medicine. Inaccurate information extraction can introduce bias and lead to inappropriate conclusions (1). Manual review, the current gold-standard for extracting information from free text, is time-consuming, tedious, and error-prone. In addition, an upper limit exists in how many EHRs a human can assess, which hinders large-scale information extraction. Introducing an automated filtering step before the manual review can improve the efficiency and sensitivity of information extraction (2). Key-word searches, commonly used as a pre-filtering step, are a crude tool and risk excluding relevant EHRs. Rule-based programming (e.g., regular expressions) and supervised machine learning have shown improved classification performance over key-word searches (2). However, these methods are costly to develop, require large amounts of labeled data, and generalize poorly across different institutions and conditions. In addition, they require fine-tuning and retraining for new tasks, making them impractical for small observational studies (2).

A new and rapidly evolving tool for information extraction is the use of large language models (LLM), a form of unsupervised machine learning that learns to predict the next element in a text sequence after being trained on a large amount of unlabeled text (3). Early LLMs used a semi-supervised approach, where models trained on unlabeled text could be fine-tuned on labeled text for specific tasks (3). Modern LLMs can solve new tasks with few or no labeled training examples (4). However, LLMs can produce true-sounding falsehoods (hallucinations) or exhibit reasoning errors (5,6). A recent study demonstrated good performance of GPT-3.5 Turbo for extracting information from veterinary EHRs (7). However, the performance of GPT-3.5 Turbo was not compared to that of other models, the nature of errors was not explored, and the influence of hyperparameter settings and text ambiguity were not assessed (7). Identifying the strengths, weaknesses and cost of different models is important for model selection. Understanding the cause of errors and the context in which they occur is crucial for optimizing model performance, and for setting of realistic expectations.

The objective of this study was to assess the classification performance of GPT-3.5 Turbo and GPT-4 omni (GPT-4o) for identifying six clinical signs associated with feline chronic enteropathy (FCE) from EHRs. In addition, we compare the reproducibility of GPT-4o to human respondents under different conditions and investigate the relationship between human interobserver agreement and LLM errors.

# 2 Methods

## 2.1 Study design and sample size

Constructed as a retrospective cross-sectional study, the sample size was determined to estimate the sensitivity of a single test (8). The calculations accounted for a type I error rate of 5%, an acceptable margin of error of 7%, an expected sensitivity of 95% and an expected prevalence of 15%. Methods for prevalence estimation are detailed in the supplementary material (Supplementary Material 1) and prevalence estimates for each clinical sign are available in Supplementary Table S1.

## 2.2 Case material

A test set consisting of 250 EHRs was sampled from all feline visits at the Veterinary Medical Teaching Hospital (VMTH) at University of California Davis, between 1985 and 2023. EHRs without text in the 'Pertinent history field' or those used for study planning were excluded (see Supplementary Material 2). EHRs for patients already represented in the test set were also excluded and replaced with resampled EHRs. The EHRs in the test set included only the admission date, presenting complaint, and pertinent history field from the original EHRs. The EHRs were manually deidentified by redacting possibly identifying information (see Supplementary Material 3).

## 2.3 Software, data type classification, scripts, and packages

The test set EHRs were analyzed using GPT-4o and GPT-3.5 Turbo (Open AI, San Francisco, CA, USA) accessed through Microsoft Azure's Open AI Application Programming Interface (API) (Microsoft Azure Redmond, WA, USA) provided by UC Davis AggieCloud Services. The account was commissioned based on a data type classification of 'De-identified patient information (with negligible re-identification risk)', a Protection Level Classification of P2 and an Availability Level Classification of A1 in accordance with UC Davis' Information Security Policy 3 (IS-3) (request RITM0074868). IS-3 is based on security standards ISO 27001 and 27002 and supports cybersecurity compliance requirements NIST 800-171, PCI, and HIPAA. The analysis was conducted using a custom Python script that leveraged the chat-completion endpoint with API version 2024-02-01. Human respondents accessed the test set EHRs through a custom online survey (Qualtrics, Provo, UT, USA). All data analysis and statistical computations and visualizations were conducted using R programming language within the RStudio integrated development environment (9,10). The custom R scripts were supported by a range of open-source packages for data science (11,12), data import and data export (13–15), statistics (16,17), and visualization (18–20). All scripts used in this study are available at GitHub (https://github.com/ucdavis/llm_vet_records).

## 2.4 Model tasks and prompt engineering

Respondents (humans and LLMs) were asked to perform two tasks: 1) Determine whether an EHR mentioned the presence six clinical signs associated with FCE (classification task) and 2) cite pertinent sections of the EHR supporting the decision (citation task). Detailed instructions for both tasks were developed through iterative evaluation and adjustment (prompt engineering), informed by human and GPT-4o responses to the tuning set. The prompt used for LLM analysis only differed from the instructions provided to human respondents in the specific instructions to output the response in a JSON format.

### 2.4.1 Classification of presence of clinical signs

The clinical signs associated with FCE were selected based on a recent diagnostic consensus statement (21) and included decreased appetite, vomiting, weight loss, diarrhea, constipation and polyphagia. The instructions specified that a 'current', or 'recently present' clinical sign qualified as 'present' and allowed answers were 'true' or 'false'. A precise time cut-off for what was considered 'recent' was not provided as preliminary experiments indicated that such a criterion led to false negative results for intermittent signs (see Supplementary Material 4).

### 2.4.2 Citation of supporting text

To trace respondent decisions for classification error analysis, respondents were instructed to cite pertinent sections of the EHR. The instructions specified that only copy-pasted text should be provided, each text section should be enclosed in quotation marks, different portions of the text should be separated by white space, and ellipses should not be used to shorten the text.

## 2.5 EHR analysis

### 2.5.1 EHR analysis by LLM
The Azure Open AI API allows setting a 'temperature' value between 0 and 2, where a temperature of 0 produces the most likely response, while higher values prompt the model to generate more varied and creative outputs (22). The test set was analyzed at temperatures 0, 0.5 and 1, which were chosen based on initial experiments that showed high failure rates and invalid JSON formats at temperatures 1.5 and 2 (Supplementary Figure S1). Each analysis was repeated five times at each temperature setting. In addition to question responses and text citations, the time to complete and the cost were documented.

### 2.5.2 EHR analysis by humans

The test set records were analyzed by five human respondents (three veterinary students and two veterinarians). The humans were blinded to each other's responses and to the responses of the LLMs.

## 2.6 Assessment of classification performance

The majority opinion (mode) of human responses was considered the reference standard and the mode of the LLM responses was classified as either a true positive, false positive, true negative, or false negative. For each clinical sign, sensitivity (also referred to as 'recall'), specificity, positive predictive value (PPV) (also referred to as 'precision'), and negative predictive value (NPV) were calculated and reported along with 95% confidence intervals, using the 'Wilson' method (23). The F1-score (the harmonic mean of sensitivity and PPV) and balanced accuracy (the arithmetic mean of sensitivity and specificity) were computed. To summarize classification performance across clinical signs, the median and interquartile range (IQR) were reported for each performance metric. The statistical significance of differences in responses between GPT-

4o and GPT-3.5 Turbo at temperature 0, as well as between different temperature settings of GPT-4o was assessed with McNemar's chi square test with continuity correction and 1 degree of freedom.

### 2.7 Assessment of reproducibility

Reproducibility was analyzed for both human respondents and repeated runs of GPT-4o. Cohen's Kappa was calculated separately for each unique pair of respondents and averaged across human pairs and pairs of repeated GPT-4o runs at each temperature.

### 2.8 Assessment of compliance with instructions

Compliance with instructions was assessed in three main areas: adherence to output format instructions, providing a true or false response to classification questions, and following citation instructions. The responses generated by the LLMs were assessed with all three areas, while human responses were only evaluated for citation compliance.
Compliance with output format instructions was evaluated using automated checks built into the API call scripts. These checks verified whether an output was provided for each question and EHR, whether the output was correctly formatted in JSON and whether it contained all the specified fields. To assess compliance with the instructions to always provide a true or false answer for classification questions, a custom R script was used to identify any forbidden NA values. For citations, compliance was evaluated using a custom R script that verified whether the citations adhered to the specified format and compared them with the EHR text for exact matches. For citations that did not exactly match the EHR text, the assessment considered the nature of the discrepancy, whether it altered the meaning of the citation, and if so, whether this change affected the associated classification response.

### 2.10 Assessment of classification errors

All instances where the mode LLM response differed from the majority opinion (mode) of human respondents were considered errors. If discrepant responses cited the same text sections, the cause of error was assumed to be a difference in interpretation. If the citations differed, it was assumed that some respondents missed relevant sections.
   Additionally, errors were further categorized based on ambiguity in the description of the clinical sign in the EHR. Clinical signs described as resolved or historical were classified as 'temporal'. Vaguely described clinical signs were categorized as 'vague'.

## 3. Results

### 3.1 Case material

The test set consisted of 250 EHRs from cat visits occurring between 1991 and 2023 and ranging in word length from 3 to 1262. Although all feline EHRs between 1985 and 2023 were initially considered, no EHRs prior to 1991 contained text in the 'Pertinent History' section of the report and were thus excluded. The flow of EHR selection is depicted in Supplementary Figure S2.

## 3.2 Classification performance

The sensitivity, specificity, balanced accuracy, and NPV for GPT-4o at temperature 0, were compared to the majority opinion of the human responders. Each of these metrics averaged over 96% across clinical signs, regardless of temperature (Figure 1). The average PPV and F1 scores were lower (80.7% and 84.4 % respectively) due to GPT-4o errors being dominated by false positives.

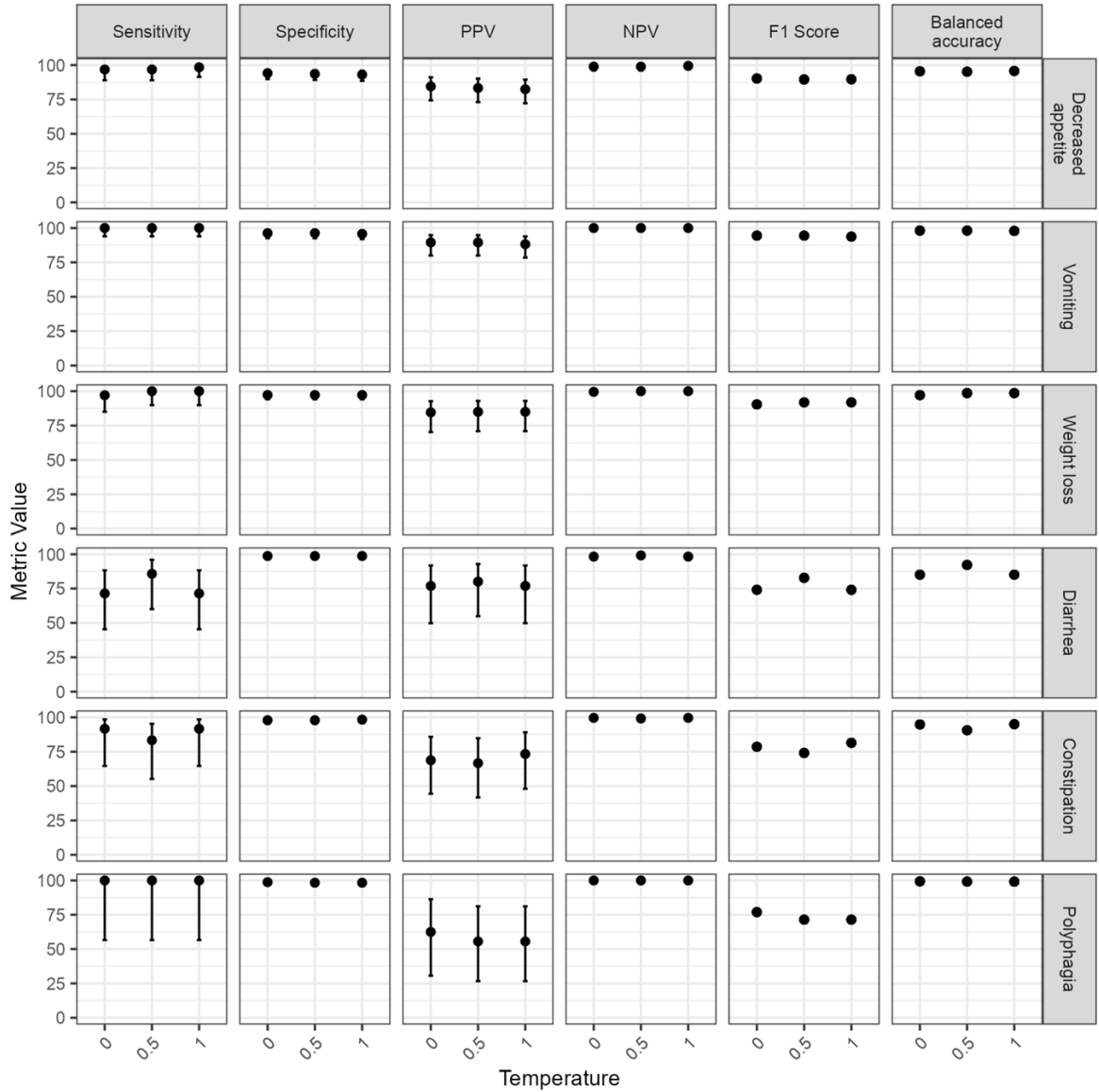

**Figure 1. Classification performance metrics of GPT-4o for extracting the presence or absence of six clinical signs at different temperatures.** Classification performance metrics for each clinical sign was computed by comparing the mode of GPT-4o responses from five repeated runs at each temperature to a reference standard composed of the majority opinion (mode) of five human respondents. Note the wide confidence intervals of classification performance metrics for three clinical signs of low prevalence in the test set (diarrhea, constipation, and weight loss) hindering interpretation of subtle variations of classification performance estimates across temperatures for these clinical signs. Error bars represent 95% confidence intervals. F1-scores and balanced accuracy are derivatives of sensitivity, specificity, and PPV; therefore, are reported without confidence intervals. NPV=negative predictive value, PPV=positive predictive value.

The temperature setting did not significantly impact the classification performance (Temperature 0 vs. temperature 0.5: $\chi^2 = 0.9$, p-value = 0.34; temperature 0 vs, temperature 1: $\chi^2 = 1.07$, p-value = 0.3, temperature 0.5 vs temperature 1: $\chi^2 = 0$, p-value = 1). GPT-3.5 Turbo performed significantly worse than GPT-4o at temperature 0 ($\chi^2 = 29.9$, p-value < 0.0001), particularly for sensitivity (81.7%, IQR 78.9-84.8%).

The average (median) performance metrics for GPT-4o and GPT-3.5 Turbo, across all clinical signs, are reported together with interquartile ranges (IQRs) in Supplementary Table S2.

### 3.3 Reproducibility

The interobserver agreement of GPT-4o responses across consecutive runs decreased at higher temperature settings yet remained higher than human interobserver agreement even at the highest temperature setting (Figure 2A). The average Cohen's kappa between repeated runs of GPT-4o ranged from 0.98 at temperature 0 to 0.93 at temperature 1, while the average Cohen's kappa between human respondents was 0.8. Supplementary Table S3 contains the summary statistics as well as Cohen's Kappa per pair of respondents for human respondents and repeated runs of GPT-4o at each temperature.

### 3.4 Compliance with instructions

Unlike preliminary experiments at temperatures 1.5 and 2.0 (Supplementary Figure S1), GPT-4o produced outputs for all questions and maintained correct JSON format for over 99% of questions at temperatures 0, 0.5 and 1.0. All output with incorrect JSON format were manually corrected. GPT-4o adhered to the instructions to provide 'true' or 'false' answers in over 99.9% of classification questions across all temperatures. However, it responded with 'NA' to one question at temperature 0 and two questions at temperature 1. Compliance with citation instructions decreased as temperatures increased for GPT-4o (Figure 2B). Human respondents complied with citation instructions less often than GPT-4o at temperature 0 but more frequently than GPT-4o at temperature 0.5 and 1. Supplementary Table S4 contains details on compliance to instructions for GPT-4o and human respondents.

Citations for both human respondents and GPT-4o at temperature 0 that did not exactly match the EHR text were manually reviewed. All discrepancies were attributed to either minor deviations in quotations, capitalization, punctation or spacing, shortening of the text, paraphrasing, or including the question or field name in the response. No hallucinations (citations not present in the EHR) were detected. Supplemental Table S5 contains details on discrepancies between citations and EHR texts.

Only one instance of a citation by GPT-4o at temperature 0 altered the meaning of the text: the EHR stated 'occasionally strains in litter box to defecate, no diarrhea' but GPT-4o shortened this to 'occasional diarrhea'. Despite this change, GPT-4o correctly classified the case as 'False' for diarrhea, indicating that the misquotation did not affect the classification outcome.

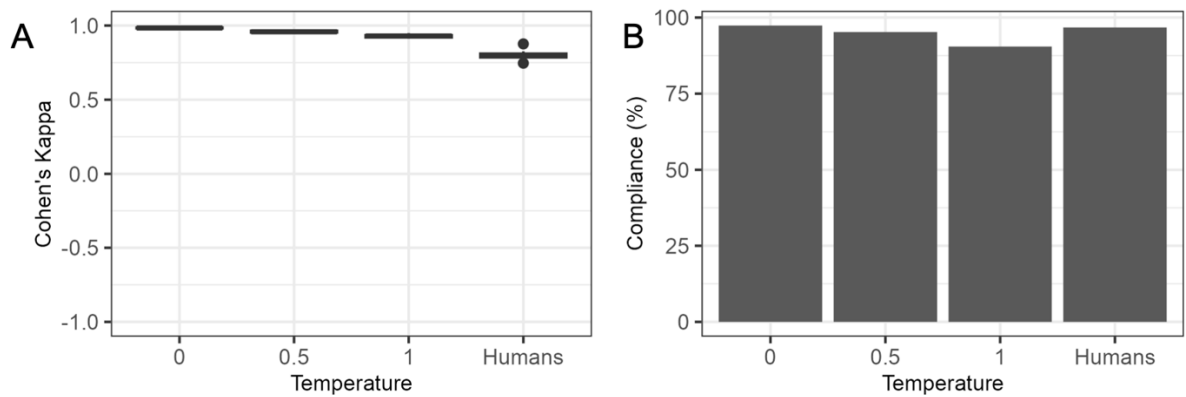

**Figure 2. Interobserver agreement and compliance with instructions for citations for humans and GPT-4o at different temperatures.** Interobserver agreement and compliance with citation instructions for GPT-4o decreased with increasing temperatures. **(A) Interobserver agreement**. Cohen's Kappa was calculated for each unique pair of human respondents, and repeated runs of GPT-4o at different temperatures. GPT-4o showed a decline in agreement between consecutive runs at higher temperatures yet maintained higher agreement than human respondents even at the highest temperature. **(B) Compliance with instructions for citations**. Compliance with instructions for citations was assessed by ensuring each citation was properly enclosed in quotes, separated by white-space and matched exactly with the EHR text. GPT-4o had slightly higher compliance than humans at temperature 0, but its compliance decreased at temperatures 0.5 and 1, falling below that of human respondents.

### 3.5 Classification errors

The mode response of repeated runs of GPT-4o at temperature 0 differed from the majority opinion response of the five human respondents in 43 out of 1,500 questions (2.9%) (Figure 3A). Most errors were false positives (35 out of 43 total errors [81.4%]). All human and GPT-4o responses at all temperatures are depicted in Supplementary Figures S3-S8).

Most GPT-4o errors (35 out of 43 total errors [81.4%]; 35 out of 1,500 questions [2.3%]) occurred in questions where at least one human assessor disagreed with the human majority, indicating potential ambiguity regarding the clinical sign in the EHR (Figure 3B). In 22 of these errors (51.2%; 22 out of 1,500 questions [1.5%]), two human respondents disagreed with the human majority opinion and sided with GPT-4o. In 13 errors (30.2%; 13 out of 1,500 questions [0.9%]), one human assessor disagreed with the majority opinion and sided with GPT-4o. Only 8 errors (18.6%; 8 out of 1500 questions [0.5%]) involved GPT-4o responding incorrectly to a question where all human respondents agreed. For instance, an EHR noting 'owner found a spit-up phenobarbital pill on the ground' led all GPT-4o runs to answer 'True' for vomiting, while all human respondents answered 'False', not citing the text section.

Classification errors where humans and GPT-4o diverged in interpretation (cited the same text but answered differently) (32/43 total errors [74.4%]; 32/1500 questions [2.1%]) were more common than classification errors where humans and GPT-4o cited different text (Figure 3B).

Classification errors with diverging interpretations of the same citation arose more frequently from EHR texts with temporal ambiguity (22 out of 32 classification errors with diverging interpretations [68.8%]; 22 out of 43 total errors [51.2%]; 22 out of 1,500 questions [1.5%]) than from EHR texts with vague descriptions of clinical signs. For example, an EHR noting a cat was 'polyphagic' at a previous visit, but 'eating less' after treatment, leading to mixed responses among human respondents. In 10 out of 32 classification errors with diverging interpretations (31.2%; 10 out of 43 total errors [23.3%]; 10 out of 1500 questions [0.7%]), the ambiguity stemmed from vague descriptions of clinical signs. For example, an EHR describing a cat with '1 bowel movement consisting of a hard, crusty piece of feces covered with a softer, outer layer' led to mixed responses among humans for constipation.

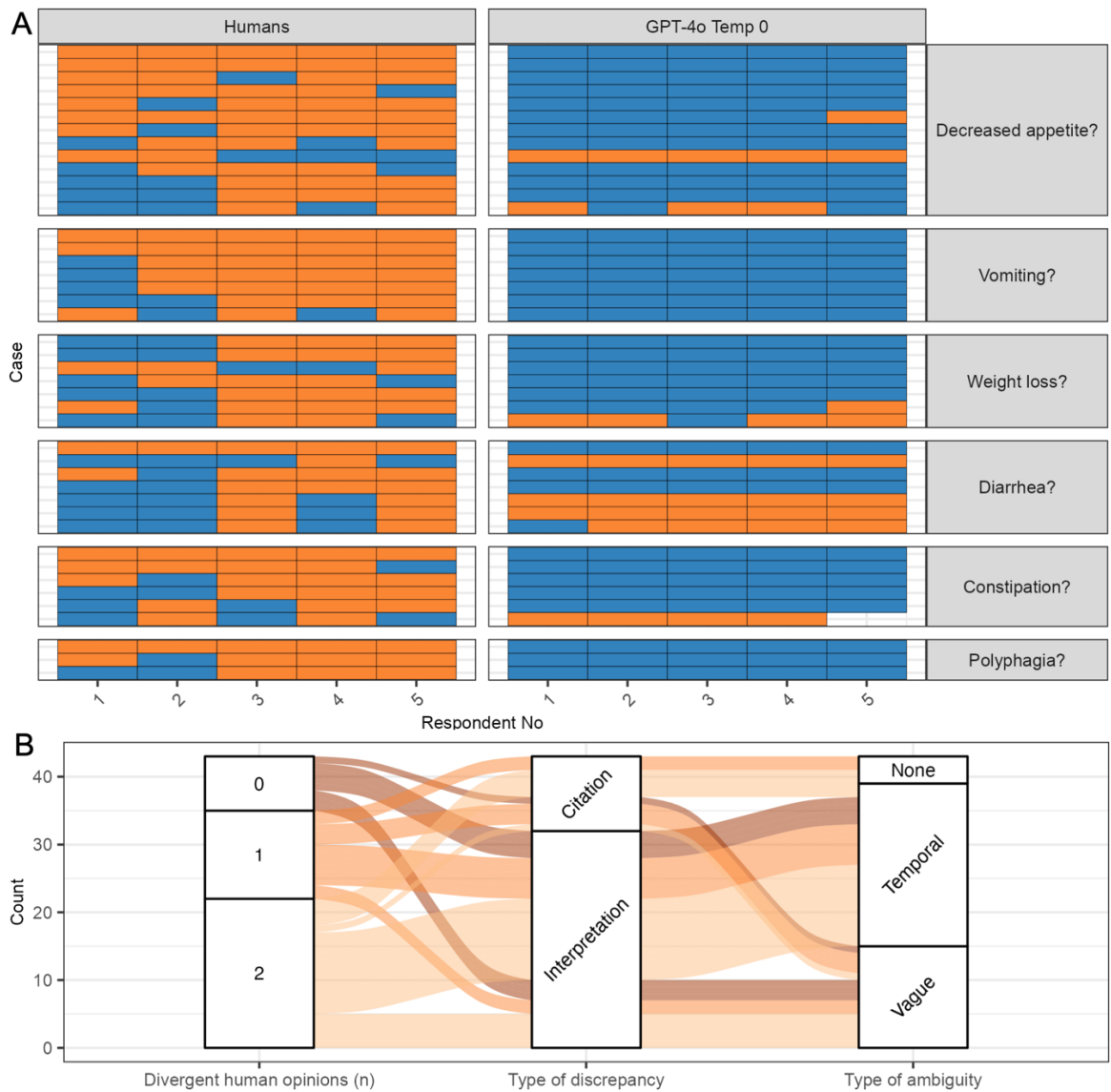

**Figure 3. Classification errors by GPT-4o at temperature 0.** All questions where the mode GPT-4o classification response disagreed with the majority opinion (mode) of human respondents were considered errors. **(A) Human and GPT-4o responses.** Five human respondents and five repeated runs of GPT-4o responded to questions on the presence of six clinical signs. False positive errors (instances where GPT-4o answered 'True' and the majority of humans answered 'False' were more common than false negative errors. Blue = True, Orange = False. **(B) Classification errors.** Most errors occurred in questions where at least one human respondent disagreed with the majority opinion. Errors with differences in interpretation were more common than errors with differences in citations. For interpretative errors, temporal ambiguity was more common than ambiguity stemming from vague descriptions of clinical signs. While some citation errors involved electronic health records with vague descriptions of clinical signs, most did not, suggesting that some respondents overlooked relevant sections of the text.

While some classification errors where humans and GPT-4o cited different text involved EHRs with vague descriptions of clinical signs, most did not, suggesting that some respondents overlooked relevant sections of the text (Figure 3B). Classification errors where humans and GPT-4o cited different text accounted for 11 out of 43 total errors (25.6%; 11 out of 1,500 questions [0.7%]), In 5 of these classification errors with diverging citations (45.5%; 5 out of 43 total errors [11.6%]; 5 out of 1500 questions [0.3%]), vague descriptions of clinical sign may have caused some respondents not to consider them as relevant. For instance, in one EHR, one human and all GPT-4o runs answered 'True' for vomiting, citing 'gagging', while 4 out of 5 humans (80%) answered 'False' without any citation. In 4 out of 11 classification errors with diverging citations (36.4%; 4 out of 43 total errors [9.3%]; 4 out of 1,500 questions [0.3%]) neither vagueness nor temporal ambiguity were appreciated, suggesting that some respondents may have overlooked a section of text. For example, an EHR noting, 'weight 6/30/01 16.5 lb, weight 11/00 20.5 lb' led 2 out of 5 humans (and all GPT-4o runs) to answer 'True' for weight loss, while 3 out of 5 humans answered 'False' without citing the text. Notably, most of these instances (3 out of 4 errors in unambiguous text [75.0%]; 3 out of 43 total errors [7.0%]; 3 out of 1,500 questions [0.2%]) involved the majority of human respondents answering 'False' while GPT-4o answered 'True', suggesting that some errors might reflect flaws in the human majority opinion rather than true GPT-4o errors. Only one instance (1 out of 43 total errors [2.3%]; 1 out of 1,500 questions [0.07%]) involved GPT-4o missing an explicit mention of a clinical sign that the majority of humans did not, indicating that GPT-4o was better at identifying relevant portions of the text then human respondents.

### 3.6 Time and cost

GPT-3.5 analysis was quicker and cheaper than GPT-4o analysis. The median time and cost per EHR were 1.6 seconds (IQR 1.4-1.9 seconds) and 0.07 US cents (IQR 0.06-0.08 cents) for GPT-3.5 and 2.5 seconds (IQR 1.9-3.3 seconds) and 0.7 US cents (IQR 0.7-0.9) for GPT-4o.

### 4 Discussion

This study demonstrated a high classification performance of the LLM GPT-4o in identifying clinical signs consistent with FCE in EHRs from a single veterinary referral hospital. The findings indicate near perfect sensitivity and negative predictive value, an outcome favorable for the intended use of the model as a screening tool. These results align with two previous studies that analyzed the classification performance of GPT models using human manual review as the reference standard. One of the studies used GPT 3.5 to identify cases of obesity in veterinary medical EHRs, reporting a sensitivity of 100% (7), while another study with a previous version of GPT-4 (1106) achieved a 97% sensitivity in identifying comorbidities in human cancer patient EHRs (24). In both these studies, as well as the current one, errors were dominated by false positives, where the LLM indicated a clinical sign as present while the human reviewer considered it absent, resulting in more modest PPVs and F1 scores.

In addition to an excellent classification performance, this study demonstrated good reproducibility, which was higher between repeated runs of GPT-4o than human respondents, regardless of temperature settings. At temperature 0, average Cohen's Kappa for repeated runs of

GPT-4o approached perfect agreement, suggesting that a single analysis run is sufficient and that averaging the results over multiple runs may not be necessary at this temperature. In contrast to previous studies, this study employed multiple human reviewers, which enabled the evaluation of interobserver agreement, and the identification of challenging records characterized by low levels of consensus among reviewers.

Most GPT-4o errors occurred in instances where human respondents disagreed, suggesting that many of these errors were 'reasonable interpretations' of ambiguous information. The majority of disagreements among human respondents involved clinical signs that were described as resolved or historical. Notably, respondents were asked to assess the 'presence' of a clinical sign and that a 'current', or 'recently present' clinical sign qualified as 'present'. Supplying a more stringent definition of 'presence', such as including time cut-offs for 'current' or 'recent', might have reduced the frequency of errors. However, this approach was ultimately dismissed to avoid the systematic exclusion of chronic intermittent clinical signs, which are vital for identifying FCE.

Although the classification performance of human reviewers cannot be assessed when using human reviewers as a reference standard, there were instances where humans missed sections of text relevant to a clinical question, resulting in discrepant responses. Similarly to a previous study (24), failing to cite explicit mentions of a clinical sign was much rarer for GPT-4o than humans suggesting that in some instances, GPT-4o may exceed human sensitivity.

Hallucinations, a major concern for implementing LLM for veterinary research, (25) were not observed in this study. However, instances of paraphrasing, shortening, and non-adherence to instructions for citations occurred even at the lowest temperature setting of 0. Although these deviations did not affect the GPT-4o response in any of the observed cases, they underscore that LLMs, unlike traditional computer programs, are probabilistic rather than rule-based, and may not always comply to the instructions provided.

In our specific setting, GPT-4o outperformed GPT-3.5 Turbo but was 10 times more expensive to run. However, it is possible that the performances of both models are more similar when applied to other tasks. For instance, Fins et al. (7) achieved near perfect sensitivity for detection of mentions of obesity with GPT-3.5 Turbo suggesting that this more cost-efficient model may be sufficient for some applications. In addition, it is conceivable that LLMs not tested in the current study perform better than GPT-4o or perform similarly at lower cost. Open-source LLMs could be particularly appealing when evaluating large numbers of EHRs, where cost is a limiting factor.

Only EHRs from a single tertiary referral clinic were included in this study. The quality of EHRs can vary across different veterinary settings, and it is unclear if our findings would generalize to less detailed records.

In contrast to ChatGPT, the online chat version of GPT, API applications through OpenAI or Microsoft Azure can be configured to comply with federal privacy regulations for human EHRs. Previous studies using LLMs for this purpose have relied on de-identified medical records (7,24), synthetic data (26) or locally run open source LLMs (26). While the de-identification of EHRs is feasible for a dataset size such as used in the current study, de-identifying larger sets of records might pose a significant barrier and negate most benefits of using an LLM over manual review. Although privacy and security regulations are less stringent in veterinary medicine than in human medicine, obtaining approval to use these types of applications is an essential point to consider prior to study execution.

In conclusion, the use of GPT-4o to extract information from veterinary EHRs can be a reliable alternative to human manual extraction. While considerations for cost and data privacy remain, this technology unlocks new possibilities for retrospective data analysis at a scale previously unattainable. Future work should focus on establishing guidelines for continuous validation and comparison of LLMs in this area, exploring the feasibility of this technology for integrating information from different institutions, and establishing frameworks for leveraging data mining in veterinary medicine at a larger scale.

## 5 Conflict of Interest

The authors declare that the research was conducted in the absence of any commercial or financial relationships that could be construed as a potential conflict of interest.

## 6 Author Contribution

JMW and SMK designed the study with contributions from JW, KLJ, and SM. KLJ, MAL, SLK, ND, and LEP performed the EHR assessment constituting the reference standard. Python code was written by JMW and JW. R code was written by JMW. The manuscript was written by JMW and SMK with contributions from the other authors.

## 7 Funding

JMW received support from the UC Davis Charles River Laboratories Digital Pathology Fellowship and the Peter C. Kennedy Endowed Fellowship for Veterinary Anatomic Pathology.

## 8 Acknowledgements

We would like to thank Danielle Harvey, Department of Public Health Science, University of California Davis, for statistical advice. Chat GPT free and Chat GPT teams, version 4o were used for manuscript and code editing.

# *Supplementary Material*

## 1  Methods for prevalence estimation

The prevalence of each clinical sign was estimated for sample size calculation using the following methods:

- Key-word searches of the electronic health record (EHR) of all feline visits at the Veterinary Medical Teaching Hospital (VMTH) at University of California Davis, between 1985 and 2023
- Manual review of a random subset of EHRs.
- Key-word searches of all EHRs containing text in the "pertinent history field".
- Manual review of the pertinent history section in a random subset of all EHRs

Supplementary Table 1 contains estimated prevalences by the different methods.

## 2  Methods for EHR sets used for study planning

The following sets of EHRs were used for study planning:

- **"Pilot free text"**: An initial pilot set consisting of 270 EHR's sampled from all feline visits at the Veterinary Medical Teaching Hospital (VMTH) at the University of California Davis, between 1985 and 2023. Records without text in any of the fields "Presenting complaint", "Pertinent History", "Physical Examination", "Problems", "Procedures", "Plans and Progress Notes", "Clinical diagnosis", "Comments", "Discharge Summary", or "Discharge Instructions" were excluded. The first 34 records in this set were manually assessed by one veterinarian (JMW), documenting the presence of clinical signs mentioned in any of the fields above in a spreadsheet and recording the time taken to assess each record. The strategy of reviewing the full records was abandoned due to the length of time needed to assess each record being deemed unfeasible for the study.
- **"Pilot history":** A pilot set consisting of 100 EHRs was sampled from records containing text in the "Pertinent history field". Text from the "Admission date", "Presenting Complaint" and "Pertinent History" fields were manually deidentified and used to construct a pilot survey in Qualtrics (Qualtrics, Provo, UT, USA). Through the survey, one veterinarian (JMW) documented the presence of clinical signs mentioned in the "Presenting Complaint field" or "Pertinent history" field, as well as the time used to assess each record.
- **"Tuning set"**: A tuning set, consisting of 10 EHR's, were selected from the "Pilot history" set and used for development of instructions (prompt engineering) and temperature settings.

## 3      Methods for de-identification of EHRs

Each EHR was assigned a consecutive case number and de-identified by manually redacting the following information:

- Patient name (replaced by "<Redacted patient name>")
- Names of other pets (replaced by "<Redacted other pet name>")
- Names of humans (replaced by "<Redacted human name>")
- Geographic locations (replaced by "<Redacted geographic location>")
- Visit numbers (replaced by "<Redacted visit number>")
- Microchip number (replaced by "<Redacted microchip number>")
- Hospital patient identification number (replaced by "<Redacted id number>")
- Other information about owner (replaced by "<Redacted other>")

## 4      Methods for development of instructions (Prompt engineering)

Detailed instructions for answering the questions and providing the reference were developed through iterative evaluation and adjustment, based on human and GPT-4o responses to 10 EHRs not used in the test set (tuning set, see Supplementary Methods for sampling of sets used for study planning). The primary focus was on refining instructions for answering true/false questions, with minor adjustments to the sections on referencing text and the example report.

The initial prompt provided the following instructions for answering true/false questions:

- "If the text indicates that the clinical sign is currently present, answer TRUE. Otherwise answer FALSE."
- "Clinical signs occurring within the previous week should be considered current even if the clinical sign is waxing and waning, and not present at the precise moment of the visit."

The tuning set records included explicit mentions of clinical signs (9), non-explicit mentions of clinical signs (2), abbreviations for clinical signs (3), mentions of intermittent clinical signs (3), mentions of historic clinical signs (2) and mentions of resolved but recent clinical signs (3). Using the initial prompt, the LLM response was satisfactory for explicit clinical signs, non-explicit mentions and abbreviations but discrepancies arose with intermittent, historic and resolved clinical signs. Increasingly stringent instructions (prompt 2-6) did not resolve this. However, less stringent instructions (prompts 7-8) led to correct answers for intermittent clinical signs, although issues remained with historic and resolved clinical signs. Given the priority of accurately capturing intermittent clinical signs, the instructions below, was selected for subsequent experiments:

- Current or recent signs: If the text indicates that the clinical sign is currently or recently present, answer TRUE.
- Recent occurrence: If the text indicates that the clinical sign occurred recently, answer TRUE, even if the record specifically states that the sign is not currently present.
- Historic signs: If the text indicates that a clinical sign was present at a historic date (not recent) and the clinical problem is either not mentioned in the current status, or mentioned as absent in the current status, answer FALSE.
- Uncertainty: If you are 50/50 on whether to answer TRUE or FALSE, answer TRUE.
- Provide an answer: Always provide an answer. Answers can be only TRUE or FALSE.

| Method | Set | Set size | Field | Decreased appetite | Vomiting | Weight loss | Diarrhea | Constipation | Polyphagia |
|---|---|---|---|---|---|---|---|---|---|
| Key-word search | Full database[3] | 183,713 | All | 6.5% | 26.3% | 11.2% | 16.6% | 2.5% | 0.9% |
| Human review[1] | Pilot free text | 34 | All | 23.5% | 35.3% | 17.6% | 20.6% | 5.9% | 2.9% |
| Key-word search | Full database[4] | 139,579 | History | 4.1% | 30.1% | 8.4% | 17.3% | 2.1% | 0.6% |
| Human review[1] | Pilot history | 100 | History | 16.0% | 23.0% | 6.0% | 14.0% | 3.0% | 4.0% |
| Human review[2] | Test set | 250 | History | 24.8% | 24.0% | 13.6% | 5.6% | 4.8% | 2.0% |

**Supplementary Table S1: Estimated prevalence of six clinical signs in feline electronic health records UC Davis 1985-2023.** Prevalence estimates were calculated for mentions of clinical signs in feline electronic health records (EHRs) using different methods, subset of EHRs, and fields of EHRs. [1]One human (JMW) [2]Five humans (majority opinion) [3]Excluding EHRs lacking free text [4]Excluding EHRs lacking history

| Model | Temp | Sensitivity Median (IQR) | Specificity Median (IQR) | PPV Median (IQR) | NPV Median (IQR) | F1 score Median (IQR) | Balanced accuracy Median (IQR) |
|---|---|---|---|---|---|---|---|
| GPT-4o | 0 | 97.0 (93.0-99.3) | 98 (96-99) | 81 (71-85) | 100 (99-100) | 85 (78-90) | 96 (95-98) |
| GPT-4o | 0.5 | 99 (89-100) | 98 (96-98) | 82 (70-85) | 100 (99-100) | 87 (76-92) | 97 (93-99) |
| GPT-4o | 1 | 99 (94-100) | 98 (96-98) | 80 (74-84) | 100 (99-100) | 86 (76-92) | 97 (95-99) |
| GPT-3.5 Turbo | 0 | 82 (79-85) | 99 (98-99) | 84 (78-92) | 99 (97-99) | 81 (79-86) | 91 (89-92) |
| GPT-3.5 Turbo | 0.5 | 82 (79-85) | 99 (98-99) | 82 (78-91) | 99 (97-99) | 82 (79-85) | 91 (89-91) |
| GPT-3.5 Turbo | 1 | 84 (81-86) | 99 (98-99) | 85 (76-94) | 99 (97-99) | 81 (80-87) | 92 (90-92) |

**Supplementary Table S2. Average large language model (LLM) performance compared to human majority opinion, across six clinical signs by model and temperature.** IQR = Interquartile range, PPV = Positive predictive value, NPV = Negative predictive value, Temp = Temperature

| Respondent | Temp | Median (IQR) | 1 vs 2 | 1 vs 3 | 1 vs 4 | 1 vs 5 | 2 vs 3 | 2 vs 4 | 2 vs 5 | 3 vs 4 | 3 vs 5 | 4 vs 5 |
|---|---|---|---|---|---|---|---|---|---|---|---|---|
| GPT-4o | 0 | 0.98 (0.98-0.99) | 0.98 | 0.99 | 0.99 | 0.98 | 0.98 | 0.99 | 0.98 | 0.99 | 0.98 | 0.98 |
| GPT-4o | 0.5 | 0.96 (0.95-0.96) | 0.96 | 0.95 | 0.95 | 0.95 | 0.96 | 0.95 | 0.96 | 0.97 | 0.96 | 0.96 |
| GPT-4o | 1 | 0.93 (0.92-0.93) | 0.93 | 0.94 | 0.91 | 0.94 | 0.92 | 0.92 | 0.93 | 0.93 | 0.95 | 0.93 |
| Humans | NA | 0.8 (0.78-0.81) | 0.88 | 0.78 | 0.80 | 0.78 | 0.79 | 0.81 | 0.75 | 0.81 | 0.83 | 0.8 |

**Supplementary Table S3: Cohen's Kappa for pairs of human respondents and repeated runs of GPT-4o at different temperatures.** IQR = Interquartile range, Temp=Temperature

| Respondent | Temp | Compliance with output format (%) | Compliance with classification instructions (%) | Compliance with citation instructions (%) |
|---|---|---|---|---|
| GPT-4o | 0 | 99.9 (7,494/7,500) | 99.99 (7,499/7,500) | 97.4 (7,302/7,500) |
| GPT-4o | 0.5 | 100 (7,500/7,500) | 100 (7,500/7,500) | 95.2 (7,142/7,500) |
| GPT-4o | 1 | 99.4 (7,452/7,500) | 99.97 (7,498/7,500) | 90.5 (6,785/7,500) |
| Humans | NA | NA | NA | 96.7 (7,255/7,500) |

**Supplementary Table S4: Compliance with instructions for human respondents and GPT-4o at different temperatures.** Frequencies of compliant responses to questions about six clinical signs in 250 electronic health records, for five repeated runs per temperature (or five human respondents). Temp=Temperature.

| Respondent | Quotation, capitalization, punctuation or spacing (%) | Shortening of the text (%) | Paraphrasing (%) | Inclusion of question or field name in the response (%) |
|---|---|---|---|---|
| GPT-4o at temperature 0 | 1.5 (113/7,500) | 0.7 (56/7,500) | 0.3 (19/7,500) | 0.1 (10/7,500) |
| Humans | 2.8 (209/7,500) | 0.1 (8/7,500) | 0 | 0.4 (27/7,500) |

**Supplementary Table S5: Discrepancies between citation and electronic health record texts.** Frequencies of different types of discrepancies between citations and electronic health records, to questions about six clinical signs in 250 electronic health records, for five repeated runs per temperature (or five human respondents).

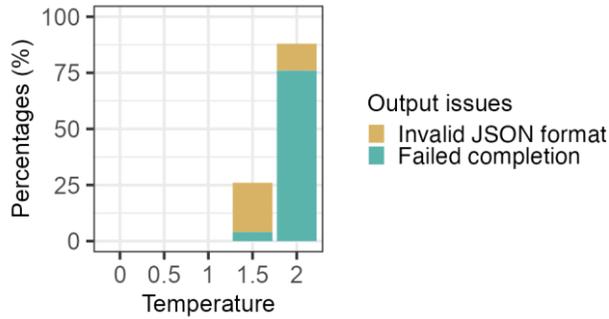

**Supplementary Figure S1: Invalid output formats in tuning set.** A preliminary experiment examined frequencies of invalid output formats for GPT-4o at temperatures between 0 and 2. High frequencies of invalid JSON format and complete output failures (due to invalid Unicode) by GPT-4o informed the decision to restrict the temperature settings examined in the test set to values between 0 and 1.

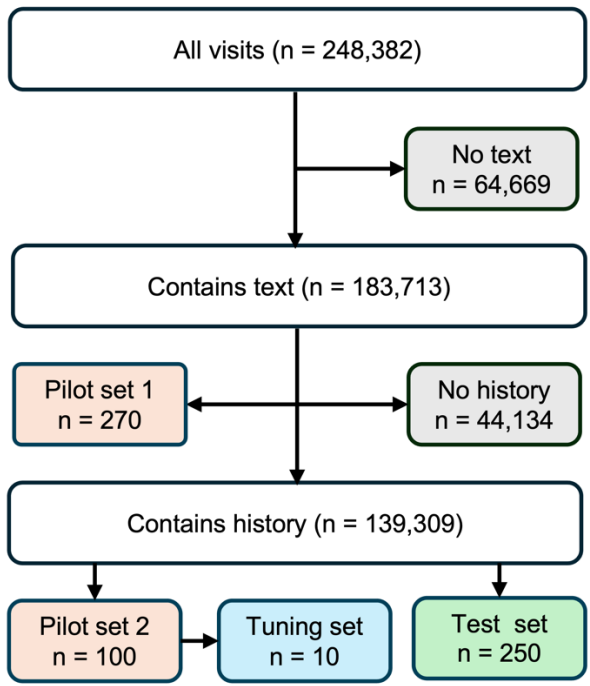

**Supplementary Figure S2: Flowchart for electronic health records.** Out of all electronic health records (EHR's) from feline visits between 1985 and 2023, EHR's for the test set were sampled from all EHRs containing history that had not been used in any pilot or tuning sets. (White = sample pools, Grey = Exclusions, Peach = Pilot sets, Blue = Tuning set, Green = Test set)

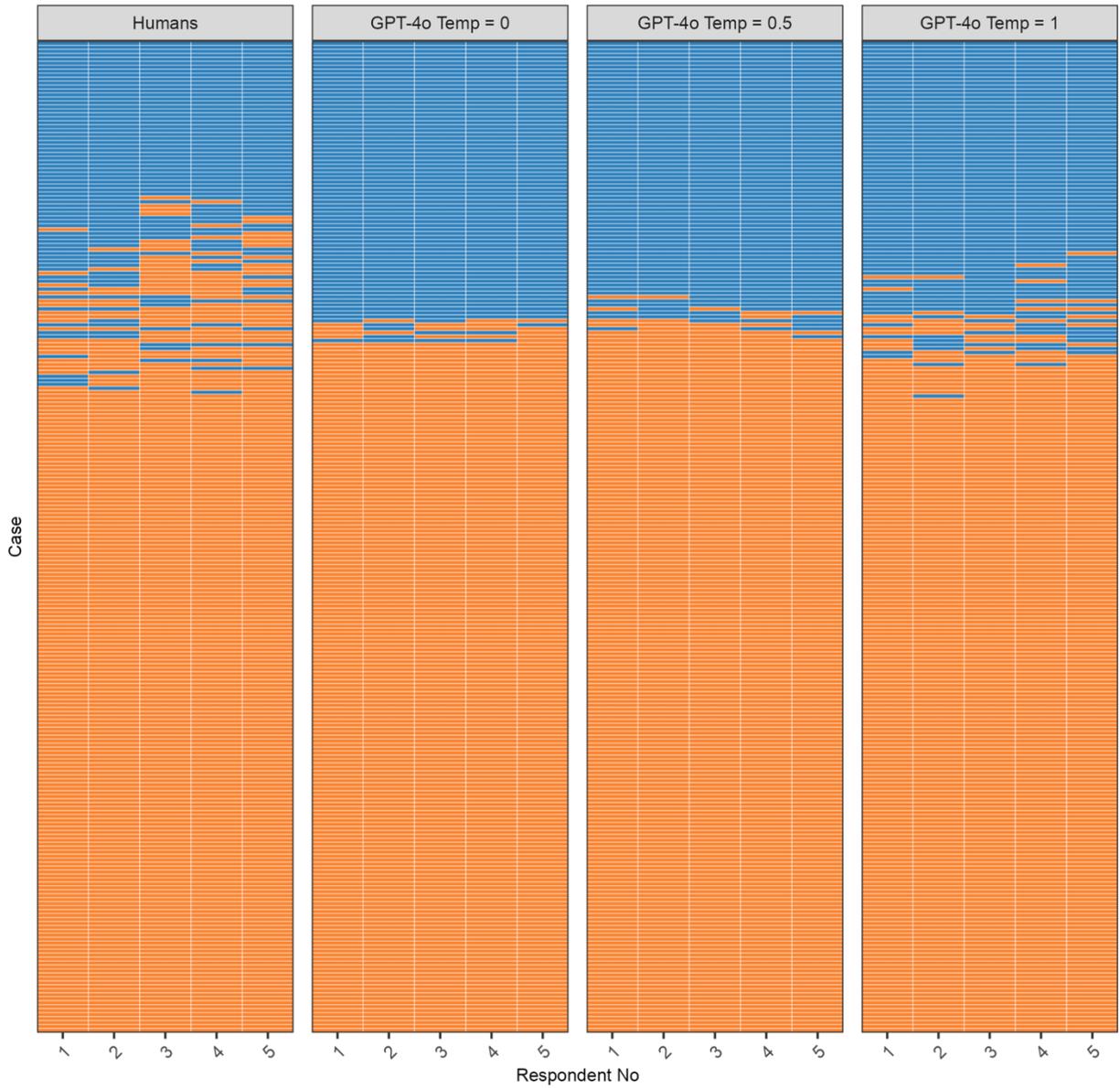

**Supplementary Figure S3: Decreased appetite? - responses for the test set.** Five humans and five repeated runs of GPT-4o per temperature setting were tasked with identifying whether each record mentioned the presence of "Decreased appetite". Blue = "True"; Orange = "False". LLM = Large language model, Temp = Temperature.

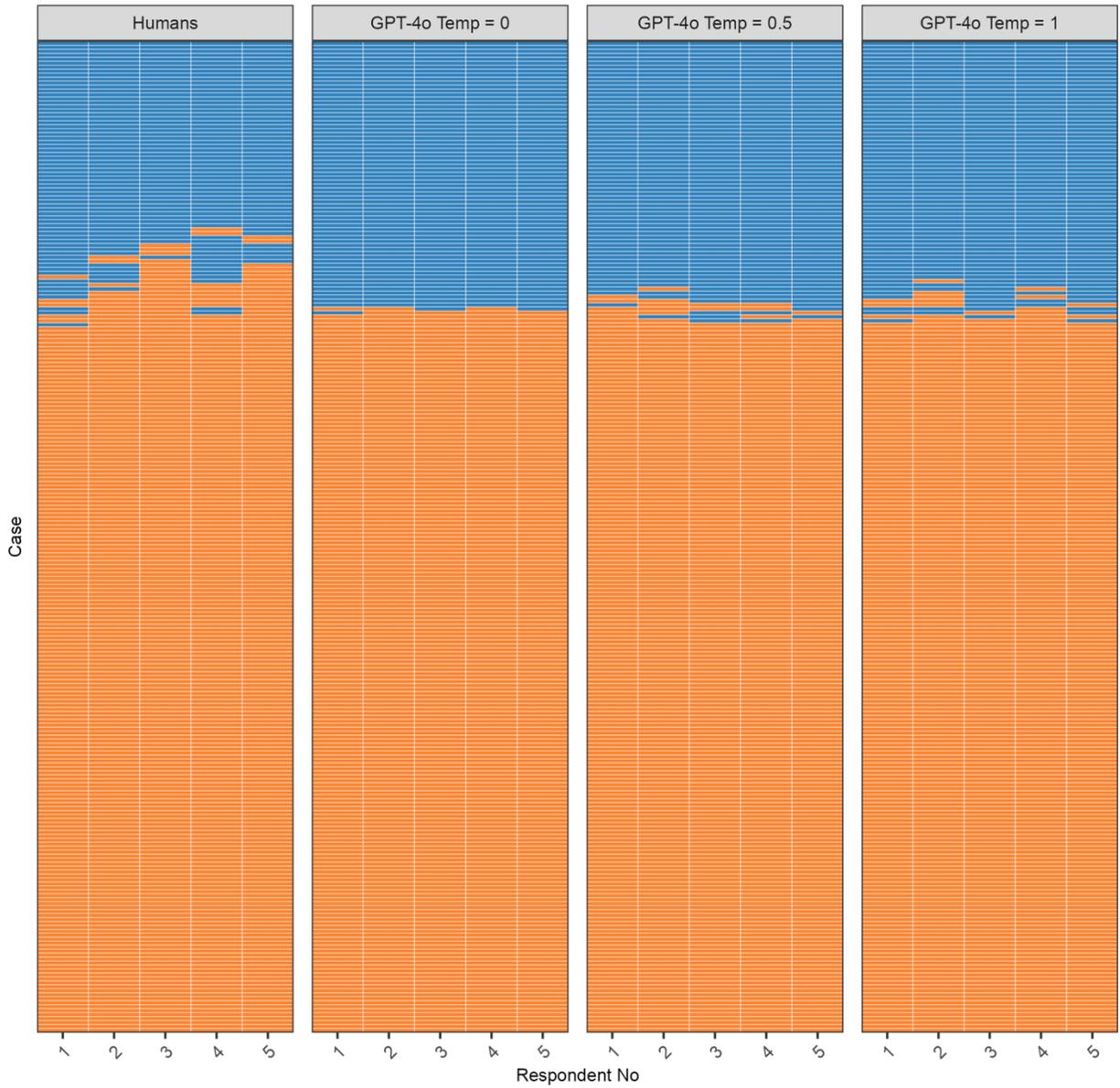

**Supplementary Figure S4: Vomiting? - responses for the test set.** Five humans and five repeated runs of GPT-4o per temperature setting were tasked with identifying whether each record mentioned the presence of "Vomiting". Blue = "True"; Orange = "False". LLM = Large language model, Temp = Temperature.

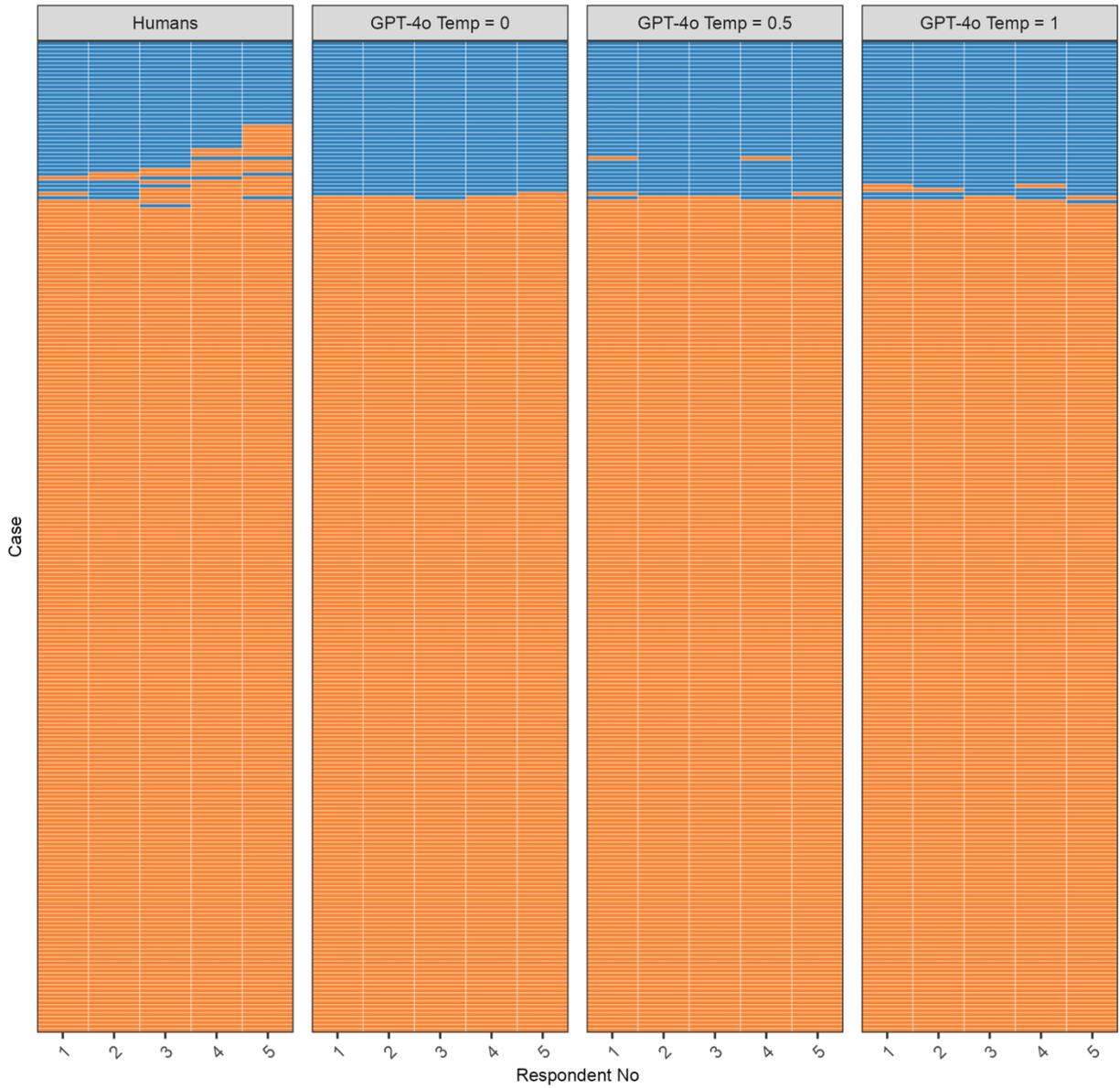

**Supplementary Figure S5: Weight loss? - responses for the test set.** Five humans and five repeated runs of GPT-4o per temperature setting were tasked with identifying whether each record mentioned the presence of "Weight loss". Blue = "True"; Orange = "False". LLM = Large language model, Temp = Temperature.

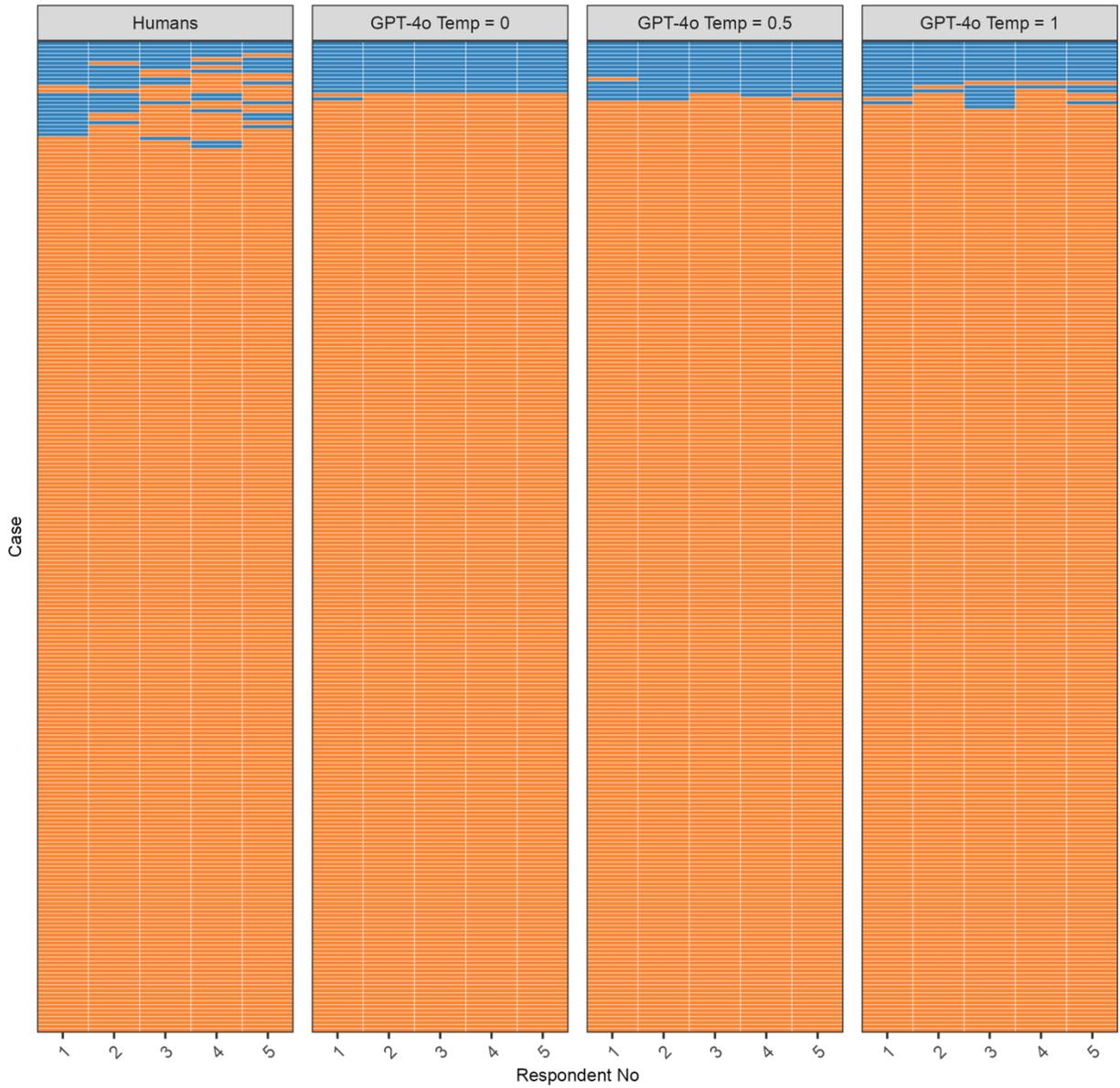

**Supplementary Figure S6: Diarrhea? - responses for the test set.** Five humans and five repeated runs of GPT-4o per temperature setting were tasked with identifying whether each record mentioned the presence of "Diarrhea". Blue = "True"; Orange = "False". LLM = Large language model, Temp = Temperature.

**Constipation?**

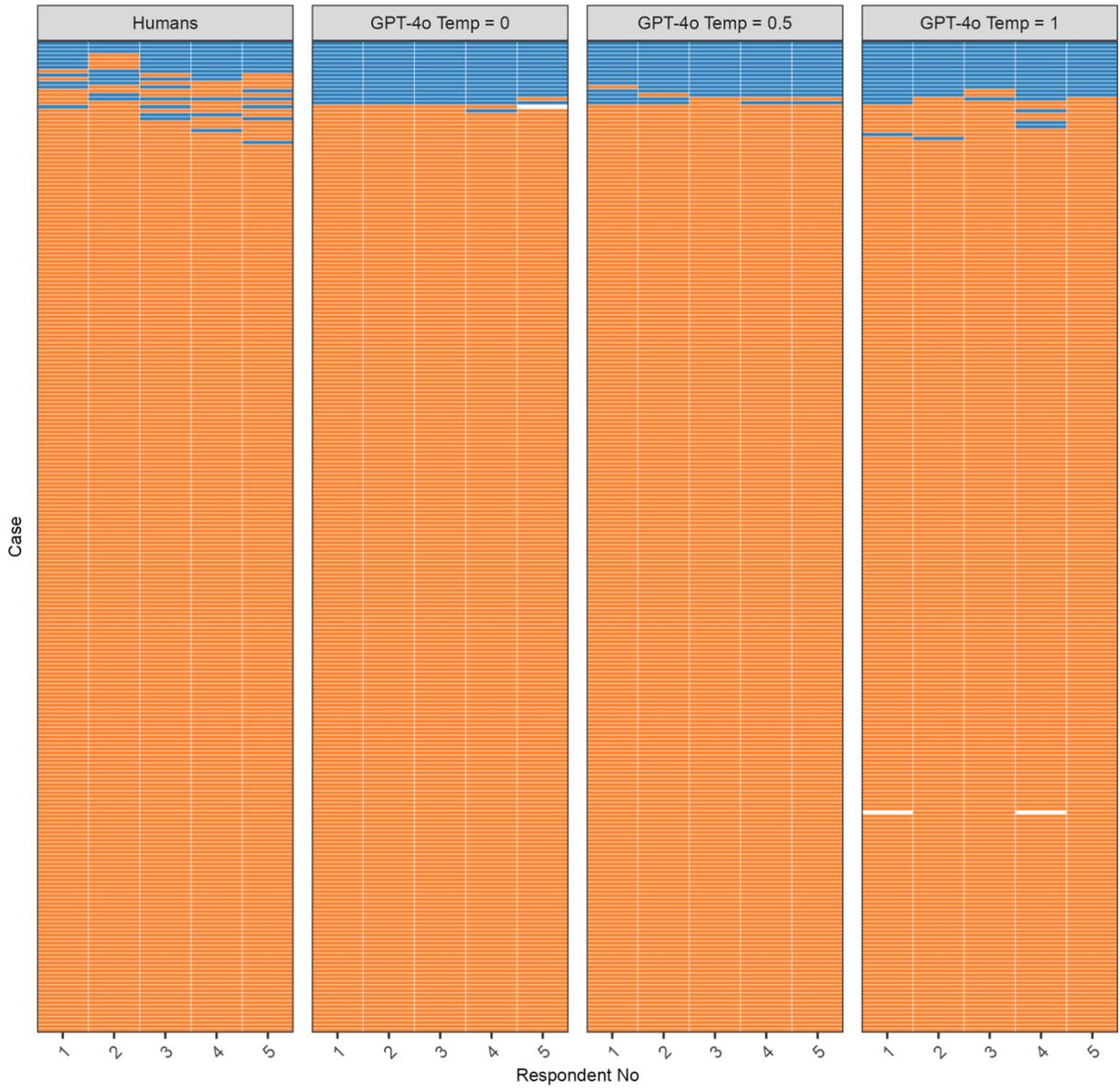

**Supplementary Figure S7: Constipation? - responses for the test set.** Five humans and five repeated runs of GPT-4o per temperature setting were tasked with identifying whether each record mentioned the presence of "Constipation". Blue = "True", Orange = "False", White = NA, LLM = Large language model, Temp = Temperature.

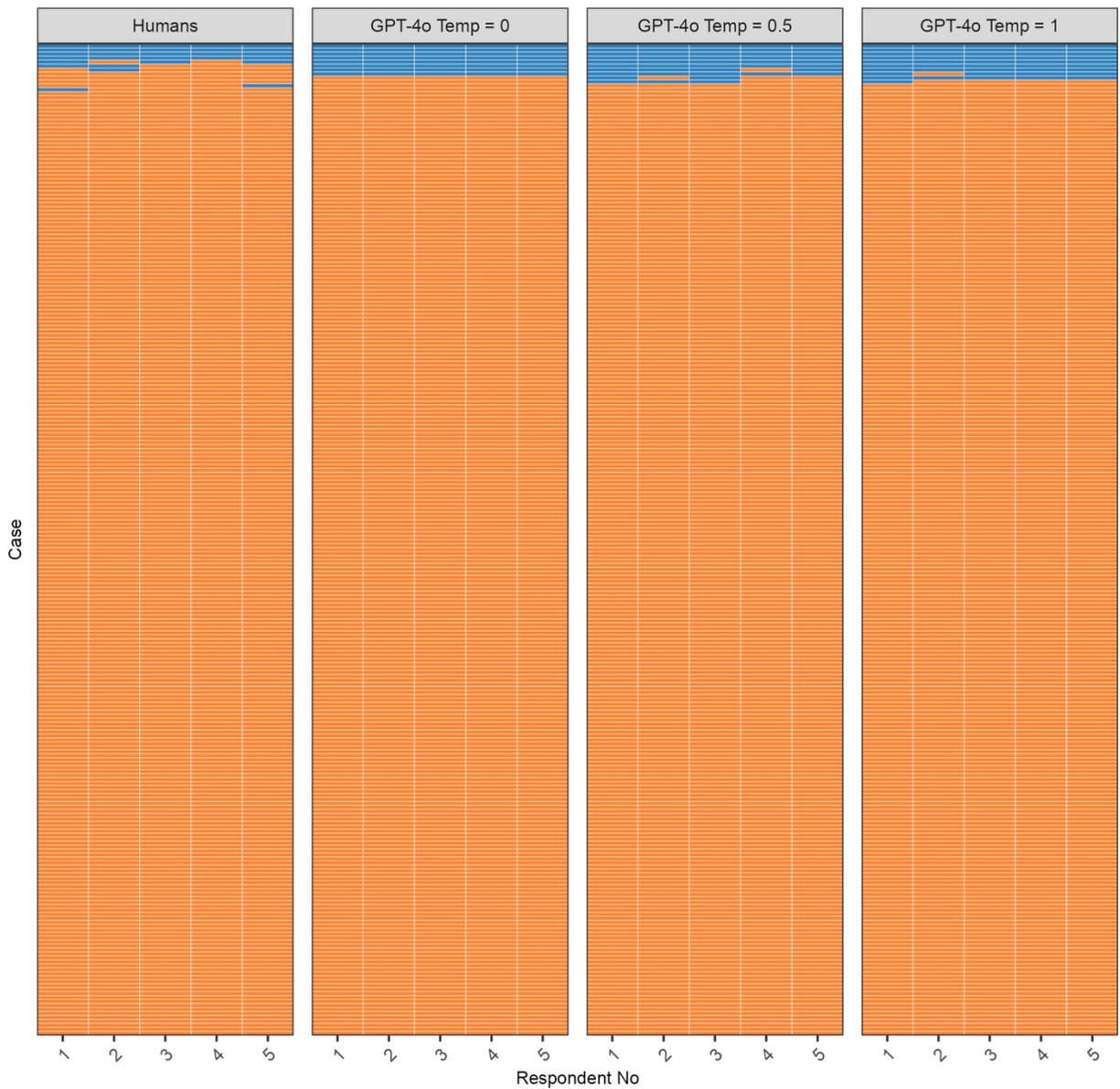

**Supplementary Figure S8: Polyphagia? - responses for the test set.** Five humans and five repeated runs of GPT-4o per temperature setting were tasked with identifying whether each record mentioned the presence of "Polyphagia". Blue = "True", Orange = "False". LLM = Large language model, Temp = Temperature